\documentclass[letterpaper, 10 pt, conference]{ieeeconf}
\usepackage[utf8]{inputenc}
\usepackage[table,xcdraw]{xcolor}
\usepackage[normalem]{ulem}
\usepackage{adjustbox}
\usepackage[ruled,vlined]{algorithm2e}
\usepackage{amsfonts}
\usepackage{bm}
\usepackage[labelfont=bf]{caption}
\usepackage{subcaption}
\usepackage{enumerate}
\let\labelindent\relax
\usepackage{enumitem}
\usepackage{amsmath}
\usepackage{float}
\usepackage{url}
\usepackage{hyperref}
\usepackage{todonotes}
\usepackage{wrapfig}
\useunder{\uline}{\ul}{}

\IEEEoverridecommandlockouts 
\providecommand{\keywords}[1]
{
  \small	
  \textbf{\textit{Keywords---}} #1
}

\title{\LARGE \bf DORA: Distributed Online Risk-Aware Explorer}
\author{David Vielfaure*, Samuel Arseneault*, Pierre-Yves Lajoie, Giovanni Beltrame}

\begin{document}

\maketitle

\begin{abstract}
    Exploration of unknown environments is an important challenge in the
    field of robotics. While a single robot can achieve this task alone,
    evidence suggests it could be accomplished more efficiently by
    groups of robots, with advantages in terms of terrain coverage as
    well as robustness to failures. Exploration can be guided through
    belief maps, which provide probabilistic information about which
    part of the terrain is interesting to explore (either based on risk
    management or reward). This process can be centrally coordinated by
    building a collective belief map on a common server. However,
    relying on a central processing station creates a communication
    bottleneck and single point of failure for the system. In this
    paper, we present Distributed Online Risk-Aware (DORA) Explorer, an
    exploration system that leverages decentralized information sharing
    to update a common risk belief map. DORA Explorer allows a group of
    robots to explore an unknown environment discretized as a 2D grid
    with obstacles, with high coverage while minimizing exposure to
    risk, effectively reducing robot failures.
\end{abstract}

\begin{keywords}
Swarm Robotics, Path Planning for Multiple Mobile Robots or Agents, Robot Safety
\end{keywords}

\section{Introduction}
The exploration of unknown environments is at the core of numerous
robotic applications from search-and-rescue operations
\cite{matos2016multiple} to space
missions~\cite{fong2005interaction}. The problem has been mostly
studied in single robot setups, but the ability to perform exploration
with teams of robots opens the door to even more ambitious
applications, because with proper coordination, the time required to
explore a given environment should decrease proportionally to the
number of robots~\cite{burgard2005coordinated}. Therefore, multi-robot
exploration is an attractive solution to many time-critical
applications such as search-and-rescue operations or planetary
exploration. Moreover, multi-robot teams are usually resilient to some
amount of robot
failures~\cite{ramachandran2019resilience,wehbe2021probabilistic,winfield2006safety}. However,
robot failures are still undesirable as they can affect team
performance and should therefore be avoided, which is the main
motivation for this work, in which we present a risk-aware exploration
algorithm for multi-robot systems: Distributed Online Risk-Aware
(DORA) Explorer.

Multi-robot systems come with their own sets of constraints and
challenges: among those, coordination and communication are the most
relevant to the exploration problem. Without coordination, the robots
will inevitably explore overlapping parts of the environment, leading
to little gains in terms of efficiency compared to single-robot
solutions. While the coordination could be optimally orchestrated from
a central computing station, such a solution would require a perfect
connectivity maintenance with each robot and a high communication
bandwidth since the robots would need to send their observations and
receive their commands. This motivates the need for a decentralized
exploration algorithm relying only on local computation onboard the
robots and communication with their neighbours.

To the best of our knowledge, there exists no risk-aware collaborative
exploration algorithm that relies solely on local or shared
information. Therefore, in this paper, we make the following
contribution to the field of multi-robot exploration: \textit{A
  decentralized exploration algorithm leveraging distributed belief
  maps (DBMs) to maximize coverage and decrease robot failure
  probability using risk-awareness.} To
evaluate this system, we test it on the specific problem of
\emph{hazard mapping} in a 2D world discretized as a grid, in which a
multi-robot team simultaneously explores a dangerous environment and
collaborates to avoid hazardous locations as well as obstacles. We
first validate our approach in a physics-based simulator, ARGoS
\cite{Pinciroli:SI2012}, in which we define a grid-based environment
with multiple radiation sources and we then test it on physical
robots. In our experiments, the belief of each cell models the
likelihood of robot failure at that point in space due to ionizing
radiation. It is worth noting that the risk of failure could originate
from any other danger type, such as fire, rough terrain, etc.

\section{Related Work and Background}
Distributed information sharing is not trivial, especially considering
the challenges of consistency and partial connectivity among the
robotic teams~\cite{amigoni2017multirobot}. The virtual stigmergy
presented in \cite{pinciroliTuple2016} and implemented in the Buzz
programming language \cite{pinciroliBuzz2016} achieves consensus among
a group of robots using conflict-free replicated data types (CRDTs),
represented as key-value pairs. This sort of shared data structure is
particularly relevant for belief maps, since it is easy to assign a
unique key to each cell based on its location.  In the virtual
stigmergy, data is shared on writing and reading the CRDT, with the
additional updates on read improving the robustness to temporary
disconnections and message drops. This solution differs from
distributed hash tables, which require a complete view of the system
at every point in time. Other distributed data storage approaches such
as SwarmMesh \cite{majcherczykSwarmmesh2020} store data in different
locations based on a fitness function instead of replicating them on
all robots. This allows the storage of more data with less
communication, but robots are less likely to have access to the latest
values.

Belief maps are a simple yet powerful tool for robotic exploration
because they can represent an environment with a 2D cell grid. They
are a generalization of occupancy maps: instead of storing only one
bit per cell to indicate the presence of an obstacle/danger, they
store obstacle/danger likelihoods and offer significant improvements
for exploration \cite{stachnissMappingExplorationMobile2003}. In the
field of multi-robot exploration, early techniques leveraging belief
maps date back as far as twenty years
\cite{kobayashiSharingExploringInformation2002,kobayashiDeterminationExplorationTarget2003},
but they rely on a fixed grid size and are tested only with two
robots. More recent works also leverage belief maps for multi-robot
exploration. For example,
in~\cite{indelmanCooperativeMultirobotBelief2018}, the robots consider
both the current beliefs and the expected beliefs from future
observations to coordinate their exploration. Grid maps and belief
maps are also widely used to train deep reinforcement learning
exploration policies
\cite{hanGridWiseControlMultiAgent,panovGridPathPlanning2018}. Such
techniques generally achieve the best performance in simulated
environments, but are usually brittle in more realistic and noisy
scenarios.

Several path planners based on Markov Decision Processes
\cite{undurti2010online,thiebaux2016rao,xiao2020robot} take into
account risk and have useful definitions of it. However, they assume a
knowledge of the global state of the environment, which is unavailable
when exploring unknown environments. Furthermore, they are so far only
applied to single-robot systems.

Many distributed exploration strategies that maximize the amount of
covered terrain have been proposed. The first approaches to stand out
in this regard are Voronoi-based coverage control
techniques~\cite{luo2019voronoi,santos2019decentralized}. A second
method covers time-varying domains, in which points of interest in the
covered region can become more or less interesting to explore,
therefore prompting a change in the coverage function
\cite{santos2019decentralized,xu2019multi}. Another method to optimize
coverage is Frontier-Based Exploration (FBE)
\cite{yamauchi1998frontier} of which many variations have been
developed, such as those based on Particle Swarm Optimization
\cite{wang2011frontier} or the Wavefront Frontier Detector
\cite{topiwala2018frontier}. However, none of these strategies take
risk into account, which is inherent to exploration.  Therefore, the
exploration strategy implemented in this paper takes inspiration of
the multi-robot control algorithm presented in
\cite{dames2012decentralized,schwagerMultirobotControlPolicy2017}
which maximizes the information gain during exploration in the
presence of unknown hazards. However, this optimal algorithm has a
very high computational complexity. In DORA-Explorer, we introduce
approximations to lower the computation load onboard the robots which
makes it is well suited for real deployment on resource constrained
robotic platforms.

We build on those approaches and address some their shortcomings by
implementing a risk-aware exploration algorithm leveraging a DBM of
the environment which is not constrained to a fixed size.

\section{System Model}
\begin{figure*}[h]
    \centering
    \begin{subfigure}{0.30\textwidth}
         \centering
         \includegraphics[width=\textwidth]{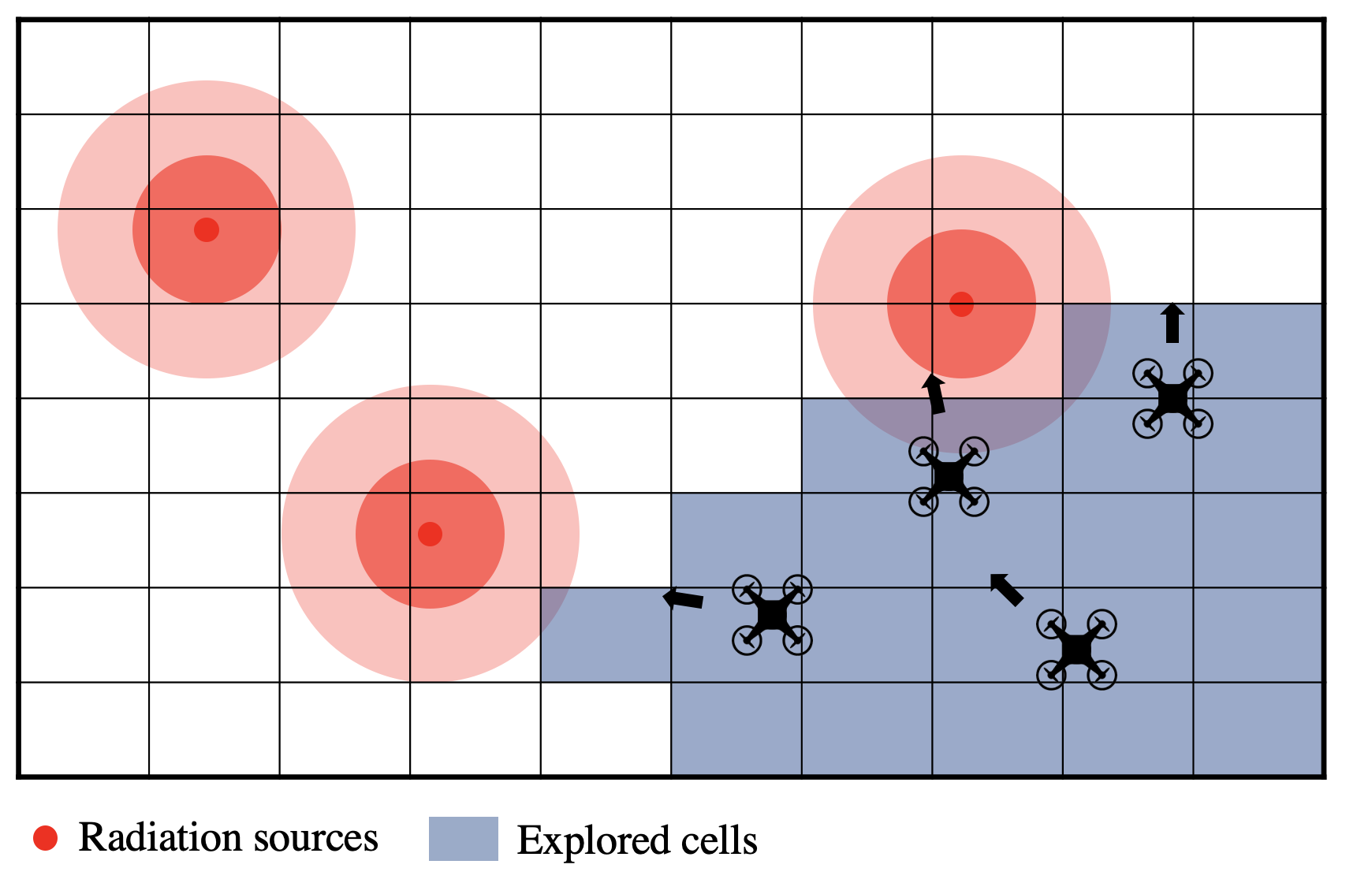}
         \caption{}
         \label{risk_aware_b}
    \end{subfigure}
    \hfill
    \begin{subfigure}{0.30\textwidth}
         \centering
         \includegraphics[width=\textwidth]{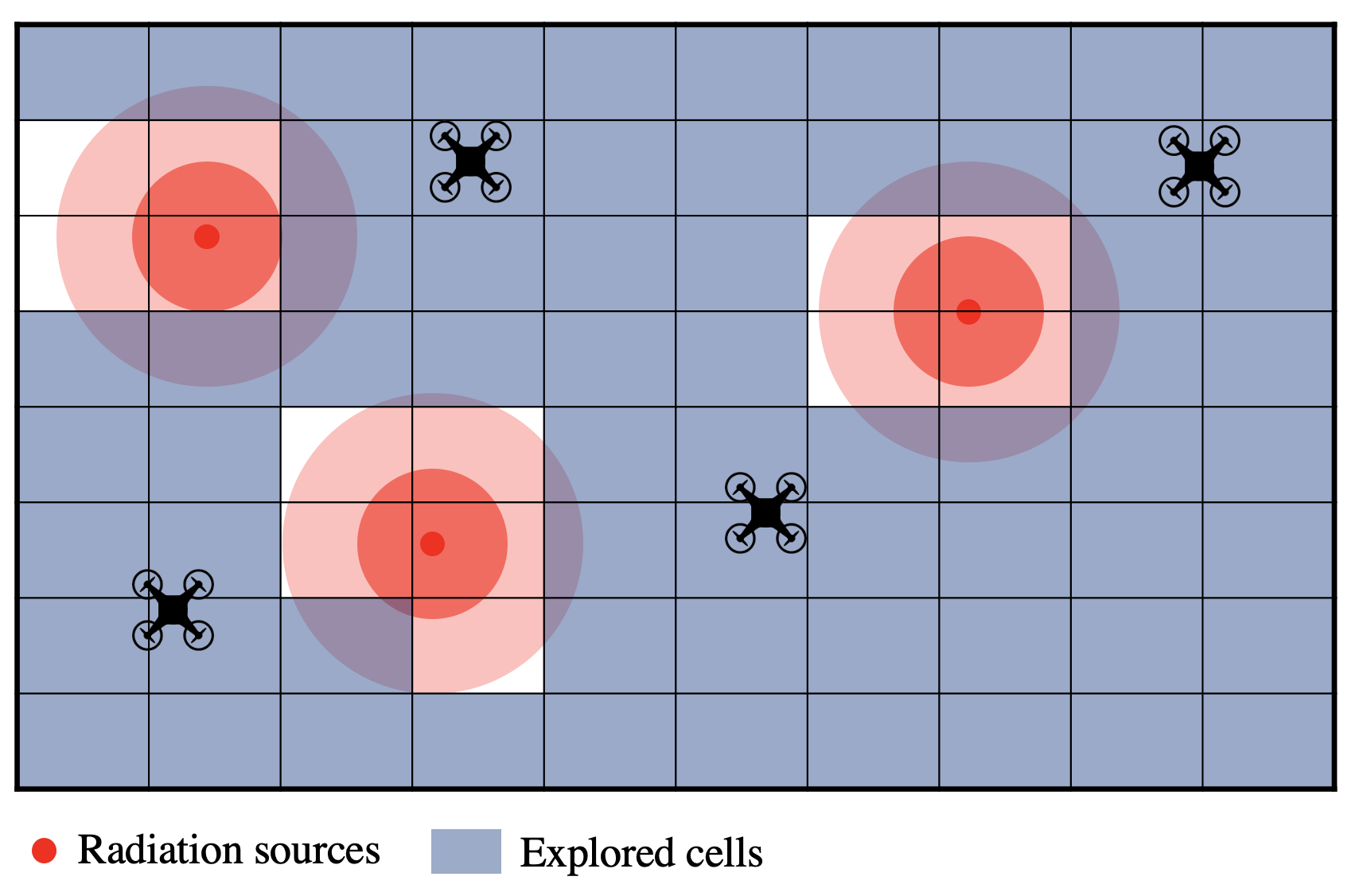}
         \caption{}
         \label{risk_aware_c}
    \end{subfigure}
    \hfill
    \begin{subfigure}{0.30\textwidth}
         \centering
         \includegraphics[width=\textwidth]{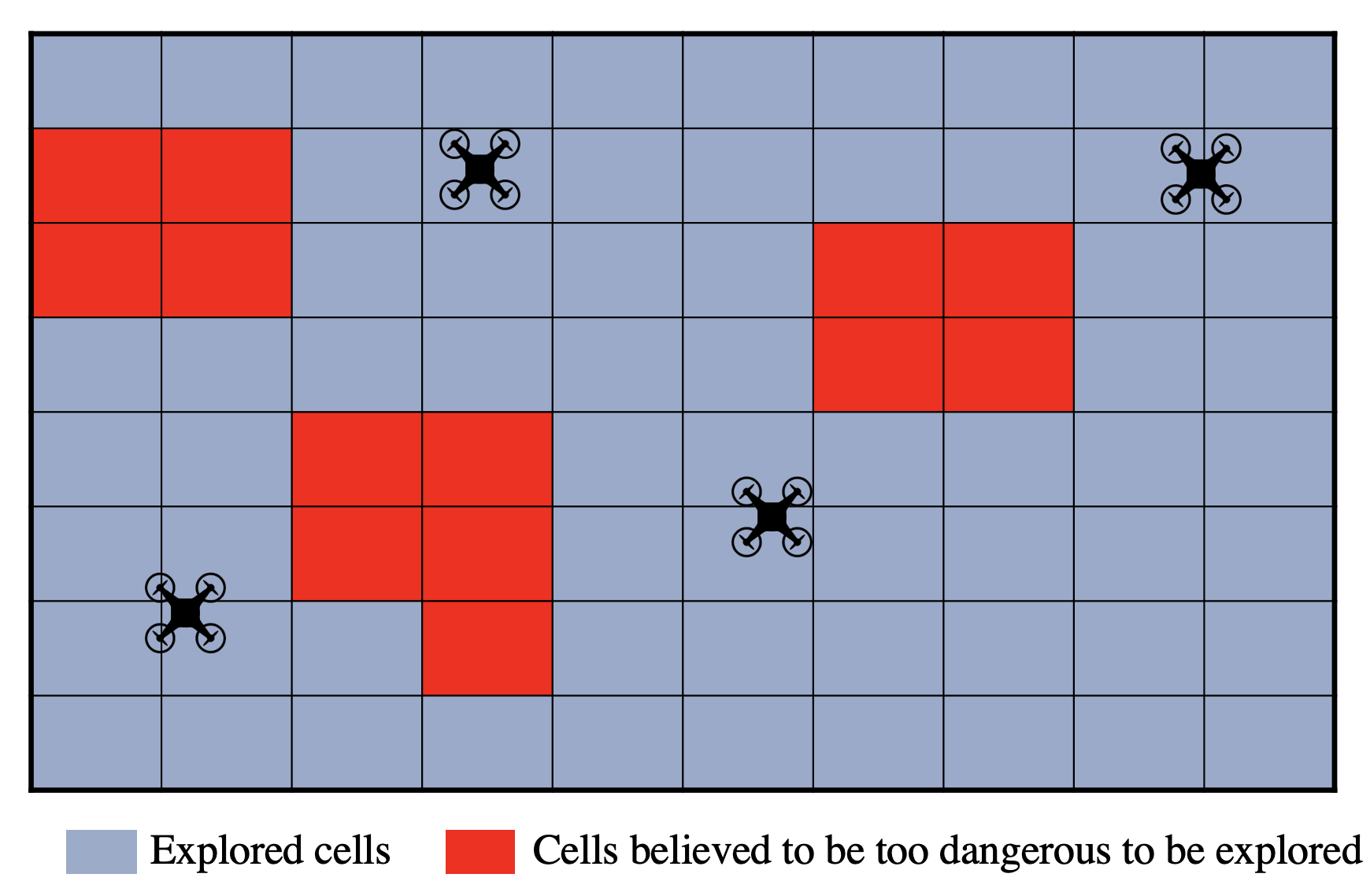}
         \caption{}
         \label{risk_aware_d}
    \end{subfigure}
        \caption{Risk aware exploration intuition. Fig. \ref{risk_aware_b}: Robots start exploring a hazardous environment. When a new cell is explored, the sensed radiation is used to update the DBM. Fig. \ref{risk_aware_c}: The cells have been mostly covered by the robots. Fig. \ref{risk_aware_d}: Only cells believed to be too dangerous remain unexplored.}
    \label{risk_aware}
\end{figure*}

The DORA Explorer leverages risk awareness to provide better
efficiency when exploring hazardous environments. Reducing the
likelihood of robot failures is of high importance as failures lead to
poor exploration performance. Indeed, if robots experiencing complete
failures are not replaced, individual failures lead to lower numbers
of robots carrying the exploration task. As a result, fewer cells are
explored at every time step and globally the rate of exploration
decreases. Avoiding dangerous areas, sometimes at the cost of not
exploring them, can be advantageous to the robot team as a whole, as
shown in Fig.~\ref{risk_aware}. By keeping robots operational, our
risk-aware exploration algorithm enables the exploration of the
environment with the contribution of all the team members. The
workforce does not decrease and as a result, the performance of the
team remains high. We model the 2D environment as cells forming a grid
represented as $E \subset \mathbb{Z}^2$. The team of robots is denoted
as the collection of agents $a_i \in A$.

\subsection{Risk Modelling}
Without loss of generality, we model risk considering point radiation
sources, denoted by the set $S$. The intensity of each radiation
source is given by $I_j\sim\mathcal{U}(0, 1)$. Each source's position
is denoted by $\bm{s}_j \in E$. Given a robot $a_i$'s discrete
position $\bm{x}_i \in E$, the perceived radiation level by that robot
from $\bm{s}_j$ is given by:

\begin{equation}
    r_{\bm{s}_j}(\bm{x}_i) = \frac{I_j}{1 + \lambda\rho^2}
    \label{eq:radiation}
\end{equation}

which decays as the distance $\rho$ between $\bm{s}_j$ and $\bm{x}_i$
increases, and $\lambda$ is a decay constant. Measurement noise is
accounted for in the form of a Gaussian background radiation
$b \sim \mathcal{N}(0, 0.05)$. The total radiation perceived by a
robot is:

\begin{equation}
    r(\bm{x}_i) = b + \sum_{\bm{s}_j \in S} r_{\bm{s}_j}(\bm{x}_i)
\end{equation}

Robots are only able to sense the radiation level associated with
their current position using an onboard sensor. They do not hold any
knowledge of where the radiations sources are located in the
environment. For the following definitions, it should be noted that
$r_{s_j}: E \rightarrow [0, 1]$ and $r: E \rightarrow [0, 1]$.  Let
the event of robot $a_i$ failing be $f_i=1$, the probability of such a
failure due to an individual source of radiation follows a Bernoulli
distribution:
$\mathbb{P}(f_i = 1 | \bm{s}_j) \sim
\mathcal{B}(r_{\bm{s}_j}(\bm{x}_i))$. We assume that the sources of
radiation affect the robots independently, so the probability of a
robot failing due to the combined effect of all radiation sources is
given by:

\begin{equation}
    \mathbb{P}(f_i = 1 | S) = \prod_{\bm{s_j} \in S} \mathbb{P}(f_i = 1 | \bm{s}_j)
    \label{eq:failure}
\end{equation}

which also follows a Bernoulli distribution such that
$\mathbb{P}(f_i = 1 | S) \sim \mathcal{B}(r(\bm{x}_i))$.

\subsection{Information Modelling}
The objective of exploring an unknown dynamic environment is to gain
information about it. Moreover, this information should be as up to
date as possible. Therefore, it is unlikely that visiting a recently
explored cell will yield any significant gain as the information
should not have changed drastically. Conversely, exploring areas
visited long ago should yield a greater information gain, and
unvisited areas should provide the highest information gain. The last
time of exploration $t_\epsilon$ by robot $a_i$ of a cell at position
$\bm{x}_i$ can be represented by the scalar field
$\epsilon(\bm{x}_i) = t_\epsilon$. Let $u_i=1$ be the event of robot
$a_i$ finding useful information in a cell and
$\Delta t = t-t_\epsilon$ the time elapsed since the cell was last
visited, with $t$ being the current time. Then, the probability of
\textit{not} finding useful information
$\mathbb{P}(u_i=0 | f_i=0, \Delta t)$ can be modelled as an
exponential distribution with the following probability density
function:

\begin{equation}
    f(\Delta t;\omega) = 
    \begin{cases}
        \omega e^{-\omega\Delta t}, & \text{if}\ \Delta t >= 0\\
        0, & \text{otherwise}
    \end{cases}
    \label{eq:information}
\end{equation}

where $\omega$ is the rate parameter of the distribution. In words,
the longer the cell has not been visited, the higher the chance
something has changed and consequently the lower the chance of not
finding useful information. It follows that the probability of finding
useful information is:

\begin{equation}
    \mathbb{P}(u_i=1 | f_i=0, \Delta t) = 1 - \mathbb{P}(u_i=0 | f_i=0, \Delta t)
    \label{eq:usefulInformation}
\end{equation}

Intuitively, no information can be acquired by failed robots, which
can be expressed as:

\begin{equation}
    \mathbb{P}(u_i=1 | f_i=1) = 0
    \label{eq:informationFailure}
\end{equation}

\subsection{Distributed Belief Map}
As previously stated, we implement a DBM using the virtual
stigmergy~\cite{pinciroliTuple2016} from the
Buzz~\cite{pinciroliBuzz2016} programming language. Because
$r(\bm{x}_i)$ and $\epsilon(\bm{x}_i)$ are both scalar fields, they
lend themselves particularly well to being stored in a CRDT at a low
cost. At each time step, the robots store their values of
$r(\bm{x}_i)$ and $\epsilon(\bm{x}_i)$ in their respective
stigmergies.  The inputs to both fields can be used as keys (more
precisely, a concatenation of $\bm{x}_{i;x}$ and $\bm{x}_{i;y}$). This
means that the cost of storing the information for a given time step
is very low, especially as the keys consist of a few characters and
the values are floating point numbers. Storing the information into
the DBM via the virtual stigmergy allows robots to share their
observations as it is accessible by every robot in the system. Thus, a
robot visiting a cell for the first could still have information from
which to compute a good control policy if this cell was previously
visited by another robot. In the event of a collision in the stigmergy
(when robots write to the same key in the same $t$) the data from the
highest robot ID is kept. When a robot writes to a key already present
in the stigmergy (from a previous time step), the new data is merged
with an average. The stigmergies do not directly store probabilities
and thus cannot be strictly considered as belief maps
\cite{kobayashiSharingExploringInformation2002}.  However,
$r(\bm{x}_i)$ and $\epsilon(\bm{x}_i)$ vary monotonically in the same
directions as $\mathbb{P}(f_i = 1 | S)$ and
$\mathbb{P}(u_i=1 | f_i=0, \Delta t)$ respectively, so they can be
used to formulate a control law based on the optimization of these
probability functions.

\subsection{Control Law}
We assume that robots can be controlled through a position-based
control law. The best control policy should attempt to minimize
probability of failure and to maximize the information gain. While the
directions achieving these individual objectives might be at odds in
the short term, they are in fact complementary in the long term
because no information can be gained if a robot failed, as defined by
\eqref{eq:informationFailure}, which means that avoiding danger
implicitly leads to more opportunities of gaining information
\cite{schwagerMultirobotControlPolicy2017}.

\begin{figure}[h]
	\centering
    \includegraphics[width=0.95\columnwidth]{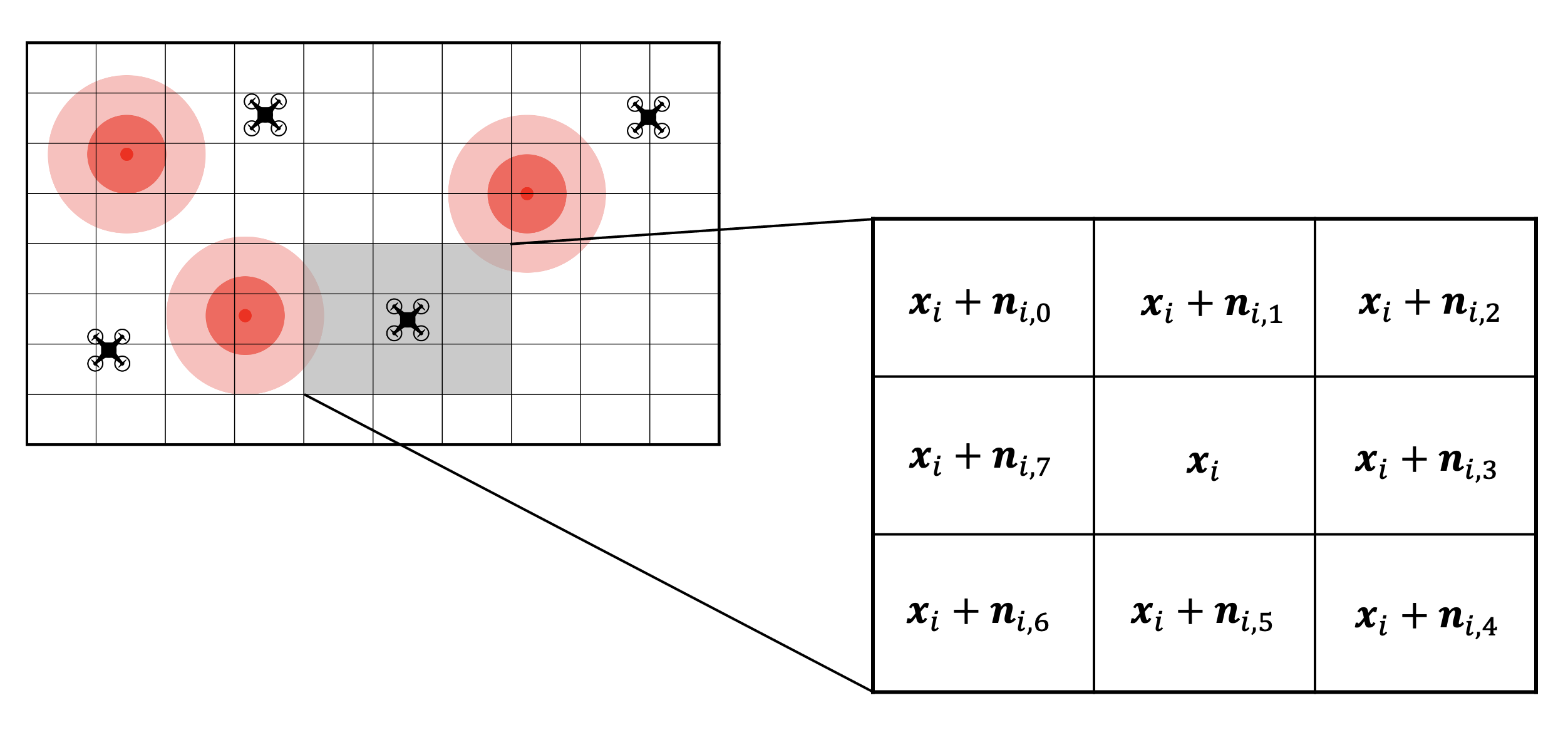}
    \caption{$\bm{x}_i$'s neighborhood. $\bm{n}_{i,0} = (-1, 1)$ is neighbor 0's offset from $\bm{x}_i$.}
    \label{neighborhood}
\end{figure}

For a robot at a given position $\bm{x}_i$, the directions where the
risk is minimized and the information gain is maximized are
respectively $\bm{\nabla}r(\bm{x}_i)$ and
$\bm{\nabla}\epsilon(\bm{x}_i)$, also denoted as $\bm{\nabla}_{r;i}$
and $\bm{\nabla}_{\epsilon;i}$. Calculating these globally at every
time step is too computationally expensive
\cite{dames2012decentralized,schwagerMultirobotControlPolicy2017}. Instead,
we compute them locally in a Moore neighborhood $\nu$ centered on
$\bm{x}_i$ as shown in Fig. \ref{neighborhood} where each neighboring
cell $\bm{n}_{i,j} \in \nu$ is a vector in $\mathbb{Z}^2$ representing
an offset from $\bm{x}_i$. We then have:

\begin{equation}
    \frac{\partial r}{\partial \bm{n}_{i,j}} = r(\bm{x}_i) - r(\bm{n}_{i,j})
    \label{eq:neighbor}
\end{equation}

\begin{equation}
    \bm{\nabla}_{r;i} = \sum_{\bm{n}_j \in \nu}\frac{\partial r}{\partial \bm{n}_{i,j}} \bm{\hat{n}}_{i,j}
    \label{eq:gradient}
\end{equation}

where $\bm{\hat{n}}$ is the unit form of $\bm{n}$. The exploration
gradient $\bm{\nabla}_{\epsilon;i}$ is calculated similarly. The
movement vector $\bm{m}_i \in \mathbb{R}^2$ for the next time step
gives a good enough approximation for short term trajectory planning
while relying only on local information and is given by:

\begin{equation}
    \bm{m}_i = \alpha\bm{\nabla}_{r;i} + \beta\bm{\nabla}_{\epsilon;i} + \gamma\bm{o}_i
    \label{eq:movement}
\end{equation}

where $\alpha, \beta, \gamma$ are respectively the risk avoidance,
exploration and obstacle avoidance control gains. The parameters can
be adjusted arbitrarily, setting them to zero will remove the effect
of the corresponding control law. The obstacle avoidance vector was
included to insure robustness and is taken from
\cite{shahriari2018lightweight}. With $\bm{\hat{m}_i}$ being the
normalized vector movement and $k$ a speed constant, the control law
for an agent $a_i$ at time step $t$ is expressed as:

\begin{equation}
    \bm{x}_i^{t+1} = \bm{x}_i^t + k \bm{\hat{m}}_i^t
    \label{eq:control}
\end{equation}

\subsection{Implementation}
The DBMs and the control law previously described are combined in
Alg. \ref{alg:dora} to describe the behavior of an agent. At every
time step, the information from the DBMs is used to determine the
agent's next movement. The DBMs are then updated with the new
information gained. Global exploration efficiency emerges through the
exchange of information through the stigmergies, but no explicit
coordination is required otherwise. Unlike FBE algorithms, DORA never
stops exploring the environment even if all cells are covered, as
information could be gained by visiting "old" cells. If
$\|\bm{\hat{m}}_i\|$ is too small, the agents move forward to avoid
stagnation.

\begin{algorithm}[h]
\small
\SetAlgoLined
\DontPrintSemicolon
 $\bm{x} \leftarrow random\_coordinates$\;
 \While{True}{
  $\nabla_r, \nabla_e \longleftarrow (0, 0), (0, 0)$\;
  \;
  \For{$n \in \nu$}{
    $\nabla_r \leftarrow \nabla_r + (r\_stig[\bm{x}] - r\_stig[\bm{n}]) \cdot normalize(\bm{n})$\;
    $\nabla_e \leftarrow \nabla_e + (e\_stig[\bm{x}] - e\_stig[\bm{n}]) \cdot normalize(\bm{n})$\;
  }
  \;
  $\bm{m} \leftarrow \alpha \cdot \nabla_r + \beta \cdot \nabla_e + \gamma \cdot compute\_avoidance(sensors)$\;
  $\bm{x} \leftarrow \bm{x} + k \cdot normalize(\bm{m})$\;
  $r\_stig[\bm{x}], e\_stig[\bm{x}] \leftarrow get\_radiation(), time()$\;
 }
 \caption{DORA Execution Loop}
 \label{alg:dora}
\end{algorithm}

\subsection{Scalability}
\label{subsec:scalability}
To achieve scalability to a high number of agents and to large
environments, DORA Explorer must have both low communication costs and
low computational costs. Lowering the communication costs associated
with sharing the belief map can be done by using the virtual
stigmergy, which is designed to limit information exchange to read or
write operations only on the requested data. Because DORA relies
solely on local information, the data transfer cost $D(A, \nu, E)$ for
an agent at a given time step is independent from the total number of
agents in $A$ and from the size of the environment $E$. For a
neighboring cell $\bm{n}_{i,j} \in \nu$, 2 stigmergy read operations
are needed per time step: one each to read $r(\bm{n}_{i,j})$ and
$\epsilon(\bm{n}_{i,j})$. In the same time step, the agent updates
$r(\bm{x}_i)$ and $\epsilon(\bm{x}_i)$ after it has moved to a new
location, which requires 2 stigmergy write operations. Each stigmergy
access requires only a few tens of bytes of data transfer for the key
and value. This mostly constant data quantity is represented as $d$,
we have $D(A, \nu, E) = 2d(|\nu| + 1)$. Similarly, the computational
cost $C(A, \nu, E)$ for the same agent at the same time step is kept
very low because of the reliance on local information only. Each time
step requires to compute 2 gradients, and referring to
\eqref{eq:neighbor}, \eqref{eq:gradient}, \eqref{eq:movement} and
\eqref{eq:control} we have that $C(A, \nu, E) = 12|\nu|+7$. The costs
related to $\bm{o}_i$ have been excluded from this analysis as
obstacle avoidance is not a critical part of DORA. The communication
and computational costs for an agent at very time step are therefore
both bounded by:

\begin{equation}
    D(A, \nu, E) \text{ and } C(A, \nu, E) \in \Theta(|\nu|)
    \label{eq:costs}
\end{equation}

Such low costs mean that DORA should scale well to a large number of
robots and should enable real time computation for even the simplest
robots.

\section{Simulations}
\subsection{Experimental setup}

We tested our system through simulations in ARGoS
\cite{Pinciroli:SI2012}, which is an open-source physics-based
simulation environment designed for robotic swarms. The agents we used
in the simulation are KheperaIV robots
\cite{kteam2021kheperaiv}. These are small round robots (140mm of
diameter) equipped with 8 infrared proximity sensors spread evenly
around their frame to perform obstacle avoidance. The
agents behavior is implemented using Buzz to facilitate swarm
management and interaction.

\begin{figure}[h]
	\centering
    \includegraphics[width=0.95\columnwidth]{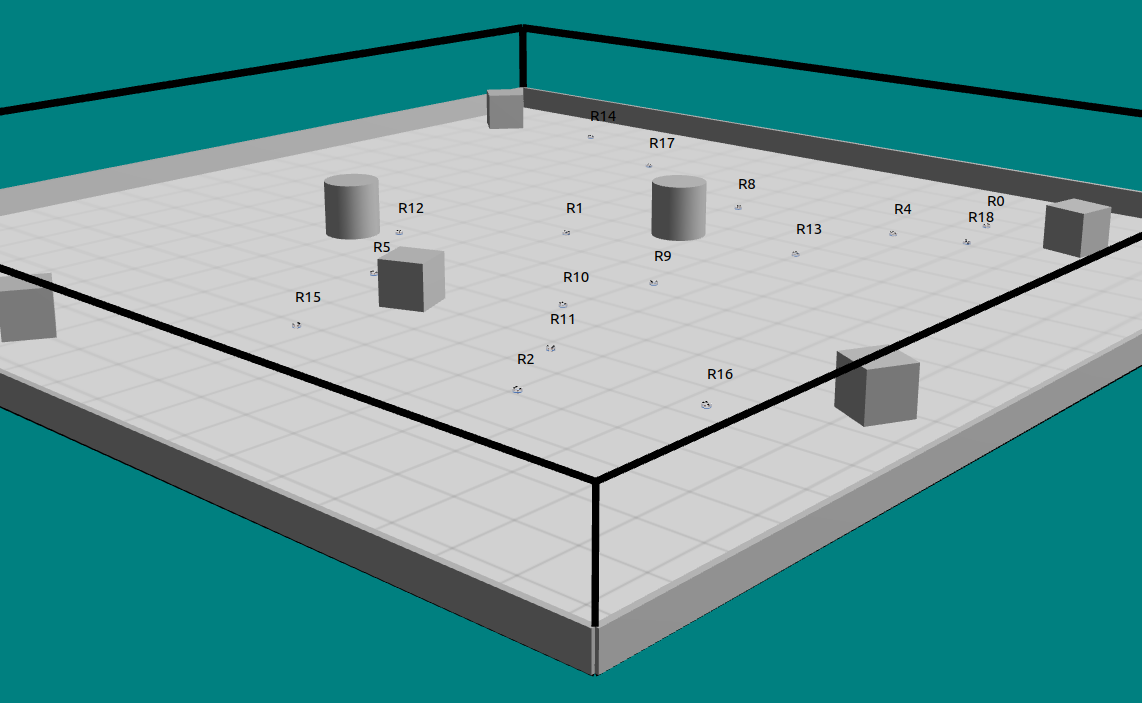}
    \caption{$400 \text{m}^2$ environment in the ARGoS simulator with 20 KheperaIV robots. Cylinders are radiation sources and boxes are random obstacles.}
    \label{argos}
\end{figure}

We deployed a set of N = \{10, 15, 20\} robots in a simulated
environment of 20m by 20m with set of 2 radiation sources. The robots'
initial positions are chosen randomly. We set $\lambda$ from
\eqref{eq:radiation} to be 5 and $k$ from \eqref{eq:control} to be
$20$. Because no information gain can be achieved by a failed robot as
seen in \eqref{eq:informationFailure}, failure must be avoided. This
leads to choosing $\alpha >= \beta$ in \eqref{eq:control}. For our
experiments, we set $\alpha=2, \beta=1$ and $\gamma=1$. The robots are
all given a random initial orientation. Radiation sensing is emulated
by an ARGoS controller reading the randomly generated radiation
sources. Failures are randomly triggered by using \eqref{eq:failure}:
if $f_i=1$, the robot stops exploring. We added 5 randomly distributed
0.8m x 0.8m obstacles to each simulation run to verify the robot's ability to
perform exploration even in cluttered environments.

We performed 50 simulation runs over 300 steps of the DORA exploration
algorithm. To assess DORA's performance, we compare it to the results
obtained by a random walk algorithm and by a FBE algorithm. The
latter's key principle is to assign one of three states (explored,
frontier, unexplored) to the cells constituting the environment and to
coordinate the robots to explore the regions near the frontier,
thereby expanding them and eventually achieving full map coverage. To
implement it, we adapted the algorithm from
\cite{yamauchi1998frontier} by having the robots share an exploration
map through a virtual stigmergy. The comparison with frontier exploration is
particularly relevant because it allows us to gain insights on our
algorithm performance in terms of terrain coverage compared to an
algorithm which was specifically designed to maximize this
objective. We also compare DORA with a random walk algorithm as a
baseline it absolutely needs to outperform. These two baselines are
commonly used for the exploration of unknown environments in the field
of swarm robotics. They do not take risk into account, but to the best
of our knowledge, no other swarm exploration strategy does.

The first metric used to assess the validity of our approach is the
number of robots which remain active (not failed) over time. This is
perhaps the most important metric because it shows how well DORA
performs in terms of risk avoidance, which is its main objective. The
second metric used to evaluate the algorithms is the total number of
cells explored by the swarm. This allows us to evaluate how well our
algorithm performs in its objective of maximizing information gain and
to verify that avoiding risk does not impact too much the exploration
performance. The third metric we studied is the communication costs of
the algorithms, measured in kilobytes of data transmitted per robot at
each time step. We included this in our analysis because it reveals if
the algorithms can scale to large number of robots.

\subsection{Results}
The following figures show an average of the results obtained in the
50 simulation runs of each algorithm.

\begin{figure*}
    \centering
    \begin{subfigure}{0.32\textwidth}
        \includegraphics[width=\textwidth]{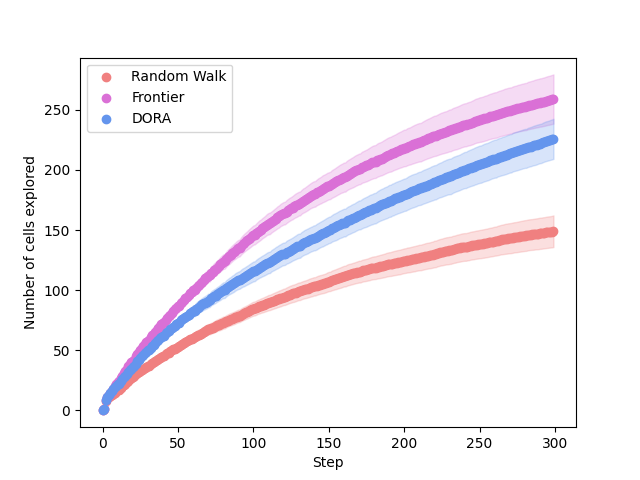}
        \caption{N=10 robots}
        \label{results:explored10}
    \end{subfigure}
    \begin{subfigure}{0.32\textwidth}
        \includegraphics[width=\textwidth]{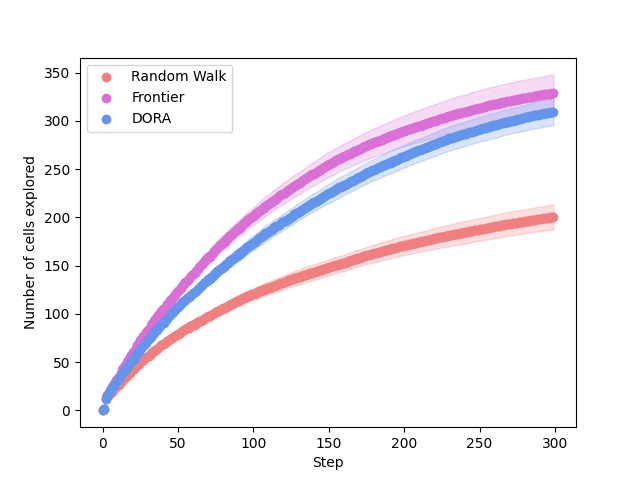}
        \caption{N=15 robots}
        \label{results:explored15}
    \end{subfigure}
    \begin{subfigure}{0.32\textwidth}
        \includegraphics[width=\textwidth]{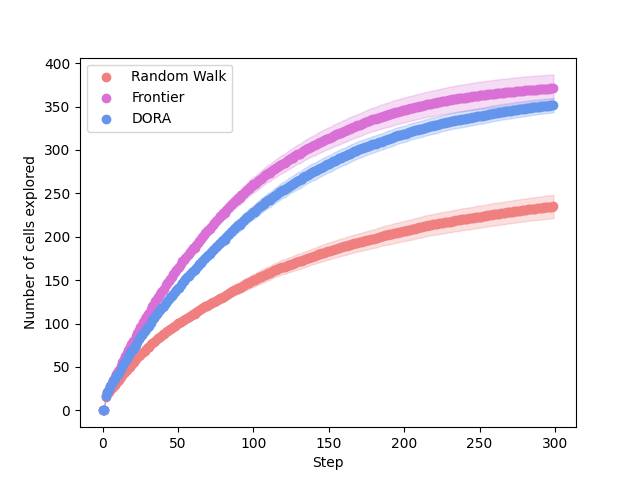}
        \caption{N=20 robots}
        \label{results:explored20}
    \end{subfigure}
    \caption{Performance comparison of DORA, FBE and random walk for number of explored cells over time.}
\end{figure*}

At the beginning of the exploration process, DORA and the baselines
perform similarly in terms of number of cells explored as shown in
Fig. \ref{results:explored10}, \ref{results:explored15} and
\ref{results:explored20}. As time progresses in the same figures, FBE
achieves a slightly higher exploration coverage than DORA, but this
gap in performance decreases as the number of robots increases. This
is an expected result, because DORA's main goal is not to achieve
maximal coverage at all costs, unlike FBE. Both FBE and DORA clearly
outperform the random walk algorithm. The other trend is that adding
more robots to the swarm results in faster exploration, with a higher
number of cells being explored for all three algorithms after 300
steps. This shows that DORA scales well to large number of robots, and
even gains in performance when swarm size increases, which is in line
with the benefits associated with swarm algorithms.

\begin{figure*}
    \centering
    \begin{subfigure}{0.32\textwidth}
        \includegraphics[width=\textwidth]{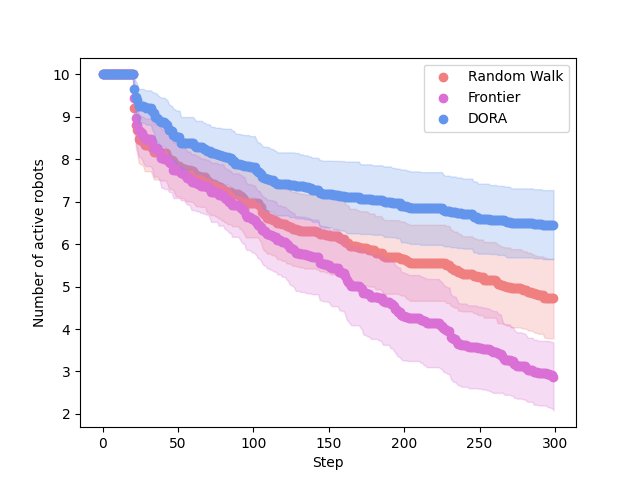}
        \caption{N=10 robots}
        \label{results:failures10}
    \end{subfigure}
    \begin{subfigure}{0.32\textwidth}
        \includegraphics[width=\textwidth]{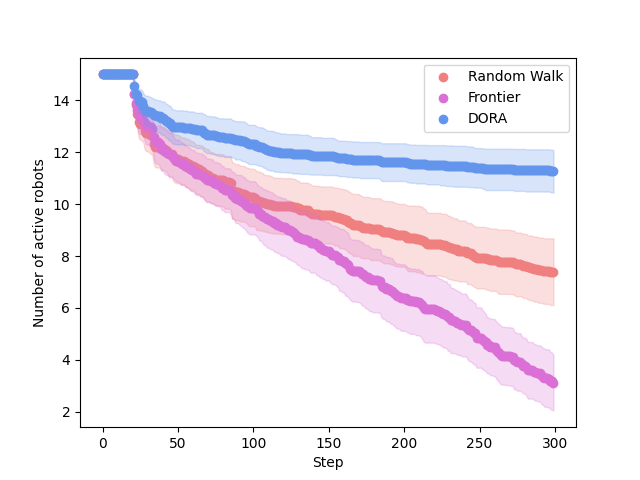}
        \caption{N=15 robots}
        \label{results:failures15}
    \end{subfigure}
    \begin{subfigure}{0.32\textwidth}
        \includegraphics[width=\textwidth]{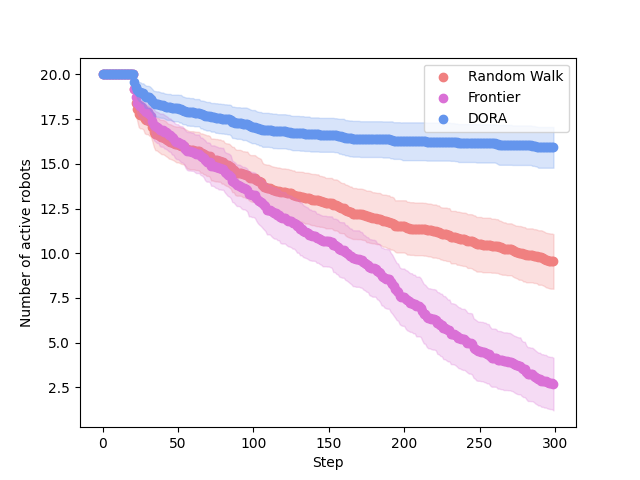}
        \caption{N=20 robots}
        \label{results:failures20}
    \end{subfigure}
    \caption{Performance comparison of DORA, FBE and random walk for number of active robots over time.}
\end{figure*}

In terms of avoiding failures, DORA unsurprisingly outperforms both
FBE and the random walk, as it is its main purpose. This is shown in
Fig. \ref{results:failures10}, \ref{results:failures15} and
\ref{results:failures20}, where DORA exhibits a higher level of active
robots over time, with this difference only increasing with larger
swarm sizes. For N = \{10, 15, 20\} robots, there are rarely any
survivors for FBE, and random walks perform only slightly better. In
contrast, DORA keeps most robots alive, achieving its objective.

\begin{figure}[h]
    \centering
    \includegraphics[width=0.95\columnwidth]{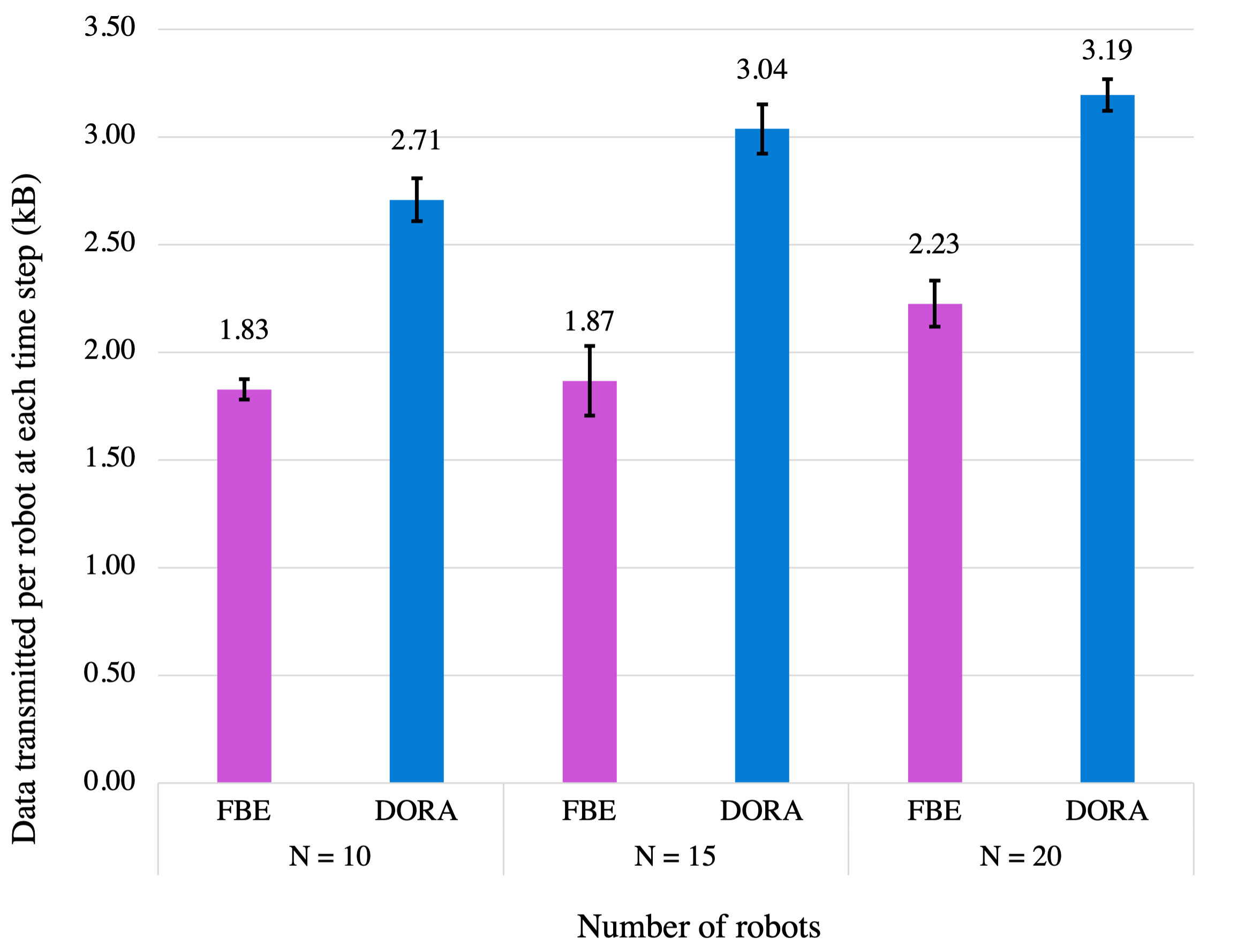}
    \caption{Communication costs for DORA and FBE}
    \label{results:communicationCosts}
\end{figure}

The results from Fig. \ref{results:communicationCosts} show the amount
of data transferred by individual agents at each time step by both
algorithms. We excluded the random walk algorithm from this figure as
it does not require any coordination or communication between its
agents. DORA transmits more data than FBE, which was expected because
the former shares information through two DBMs, while the latter uses
only one. In section \ref{subsec:scalability}, we predicted that the
amount of data transmitted at each time step would only depend on the
size of the neighborhood used, and this is confirmed by
Fig. \ref{results:communicationCosts}, where it remains roughly
constant for different number of agents. The small increase in data
transmission with increasing number of robots can be attributed to
packet collision. Overall, the amount of required data transfer is
small in comparison with the capacities of the KheperaIV robots which
communicate with the 802.11 b/g WiFi protocol.

\begin{figure*}
    \centering
    \begin{subfigure}{0.32\textwidth}
        \includegraphics[width=\textwidth]{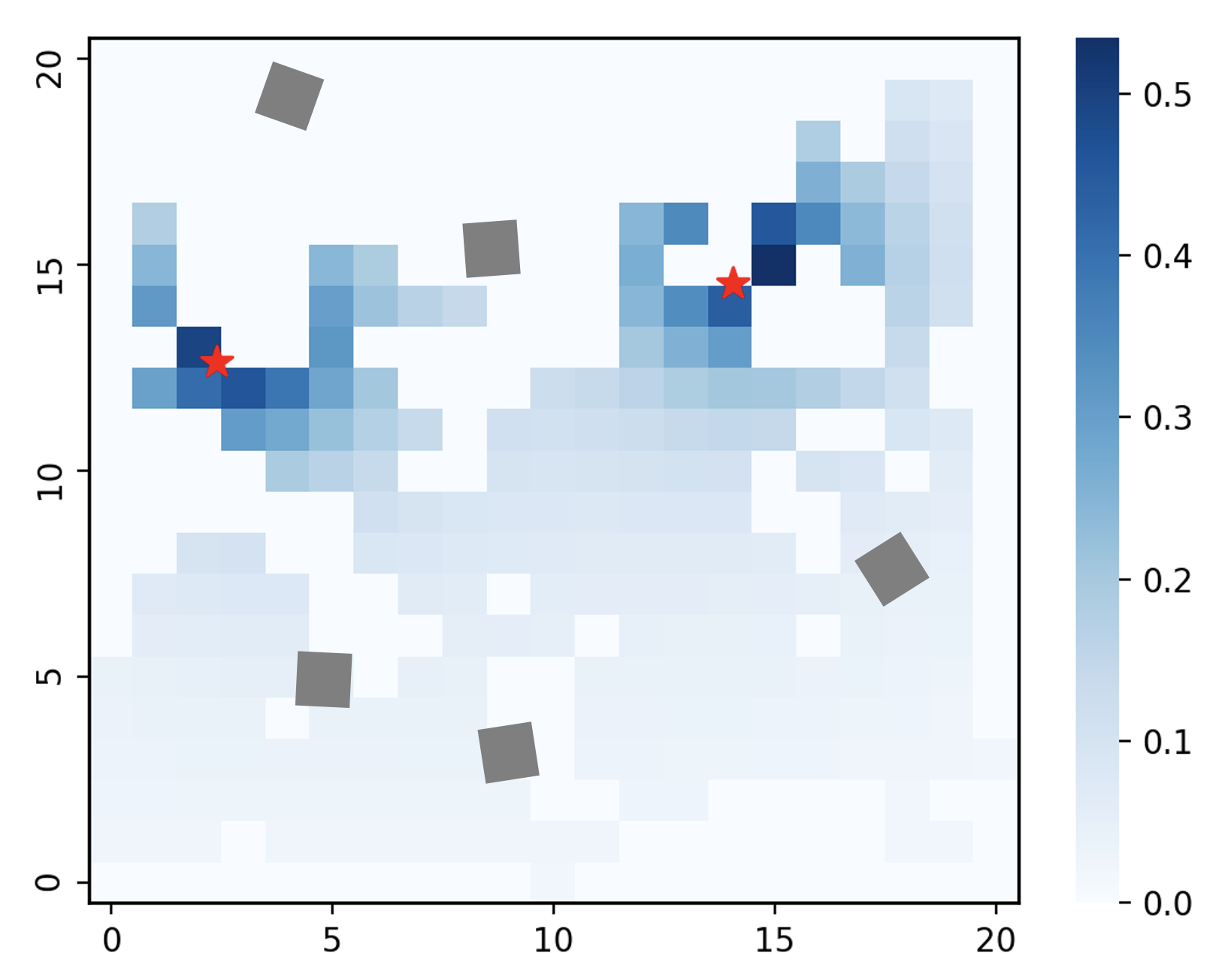}
        \caption{Random walk}
        \label{results:beliefrandom}
    \end{subfigure}
    \begin{subfigure}{0.32\textwidth}
        \includegraphics[width=\textwidth]{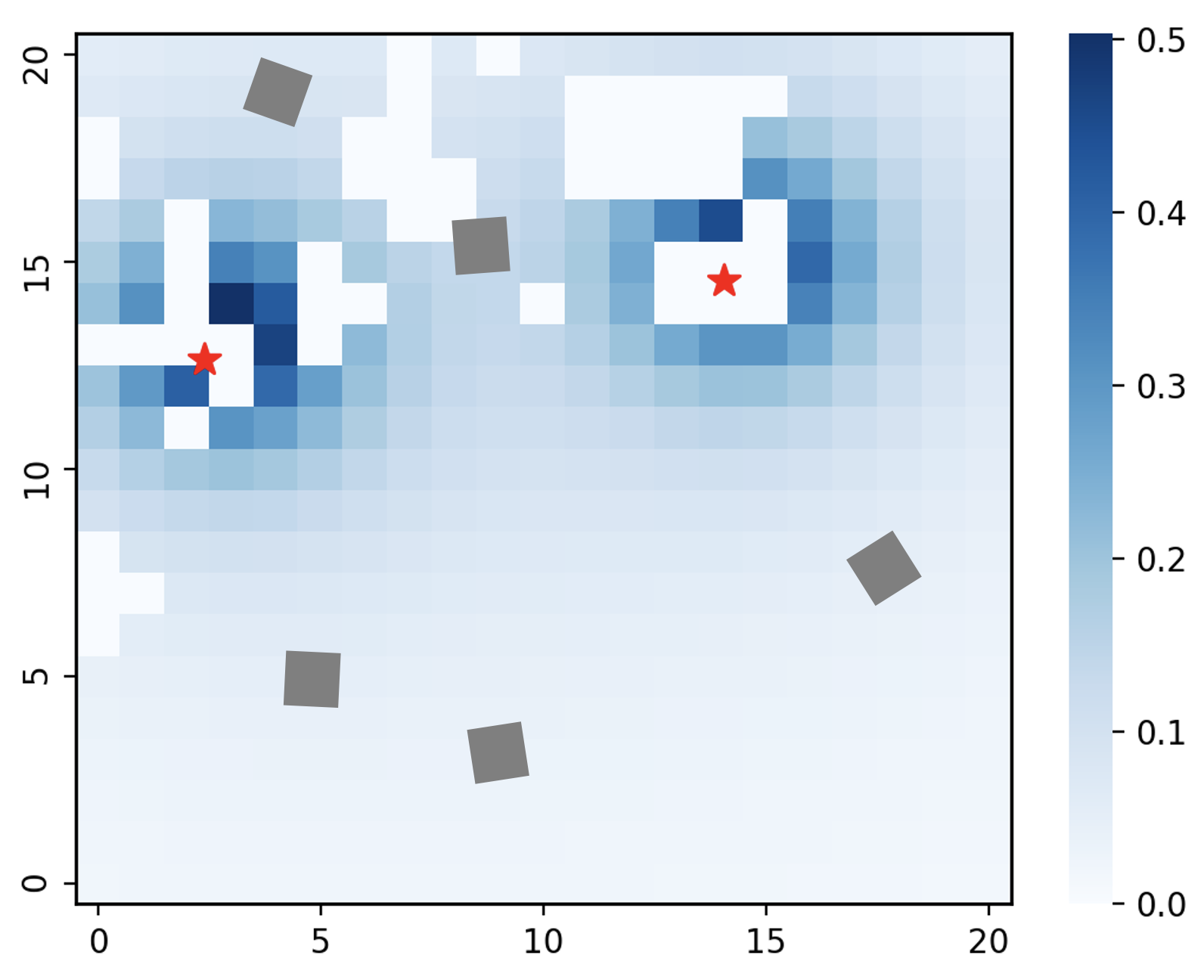}
        \caption{FBE}
        \label{results:belieffrontier}
    \end{subfigure}
    \begin{subfigure}{0.32\textwidth}
        \includegraphics[width=\textwidth]{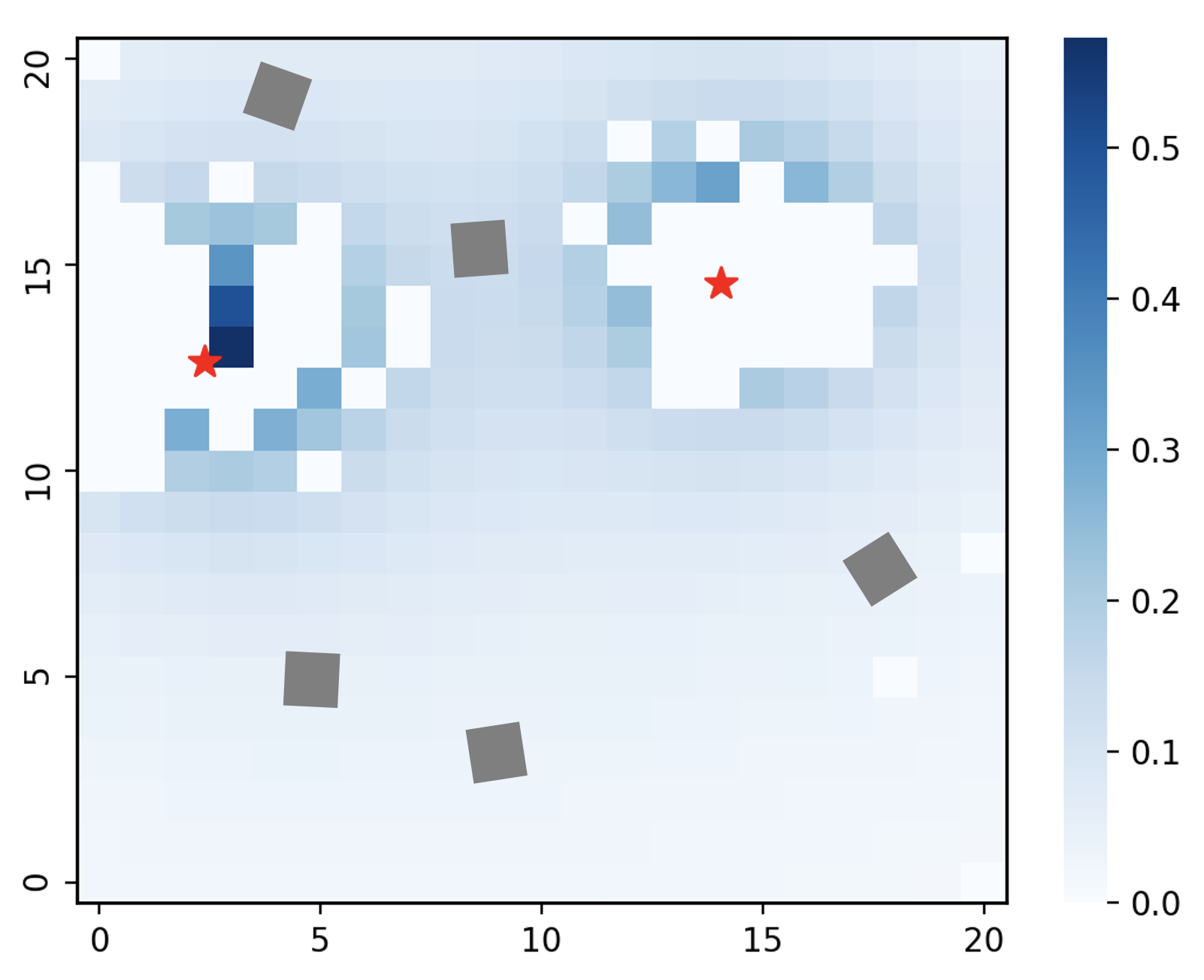}
        \caption{DORA}
        \label{results:beliefdora}
    \end{subfigure}
    \caption{Radiation belief maps of the 20m x 20m environment for each exploration algorithm of one specific simulation. Blank cells are unvisited areas, red stars are the point radiation sources and grey squares are the randomly generated obstacles.}
    \label{results:belief}
\end{figure*}

Fig. \ref{results:belief} shows the DBMs obtained at the end of an
arbitrarily selected simulation where N = 20 for each algorithm.  In
other words, it represents which cells were explored by each algorithm
and the sensed radiation intensity associated with them for one specific run. The random
walk covered much fewer cells than DORA and FBE, which both covered
roughly the same areas of the map, with the same sections remaining
unexplored. However, these areas remained unvisited for different
reasons. For FBE, the parts of the environment close to the radiation
sources remained uncovered because its agents failed when approaching
them. In contrast, DORA did not explore these cells because it
\textit{avoided them}. Again, DORA achieves very similar coverage than
FBE but does so with less robot failures. In this particular
simulation, DORA finished with 18 active robots, random walk finished
with 7, and FBE with none.

\section{Physical experiments}
\subsection{Experimental setup}
\begin{figure}[h]
    \centering
    \captionsetup{belowskip=-20pt}
    \includegraphics[width=0.65\columnwidth]{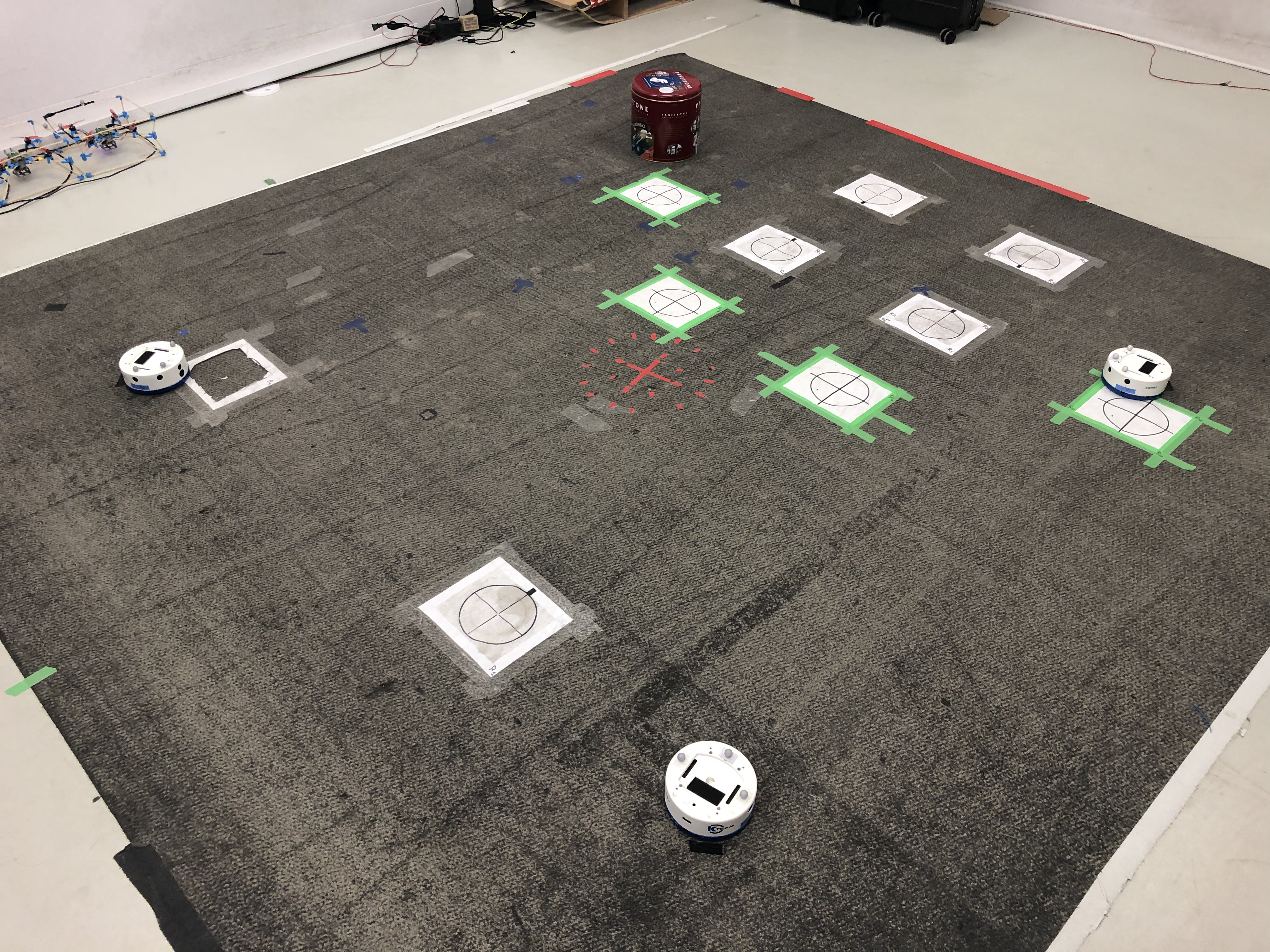}
    \caption{Experiments on three physical KheperaIV robots. The red canister represents the point radiation source in the environment.}
    \label{arena}
\end{figure}

In addition to the extensive simulations conducted in ARGoS, we tested
our system on a team of three physical KheperaIV robots in a 2m x 2m
environment containing 1 point radiation source as shown in
Fig. \ref{arena}. We only used one radiation source because of the small size of the environment.

The environment is discretized as a 10 x 10 grid, meaning that each of
the 100 cells of the grid is 20cm x 20cm large. Because the arena in
which we conducted the experiments was already limited in terms of
space we decided not to add obstacles. 
Positioning of the
robots is done using an OptiTrack motion capture system. For outdoor
applications where the environment is bigger, positioning of the
robots could instead be achieved by GPS. Radiation sensing is emulated
by an on board controller that reads the distance between the robot
and the radiation source to determine the current radiation
level. Failures are then triggered using equation \eqref{eq:failure},
in other words the higher the radiation level, the higher the
probability of failure. If a robot fails, it stops moving and as a
result stops contributing to the exploration effort. The point
radiation source is located in a corner of the arena and the robots
are initially placed in the three remaining corners. The robots'
initial orientations are chosen randomly.  We performed 5 runs over
200 steps of the DORA exploration algorithm. Again, to assess DORA's
performance, we compare it to the results obtained by FBE and random
walk algorithms.

\subsection{Results}
\begin{figure}[h]
    \centering
    \includegraphics[width=0.95\columnwidth]{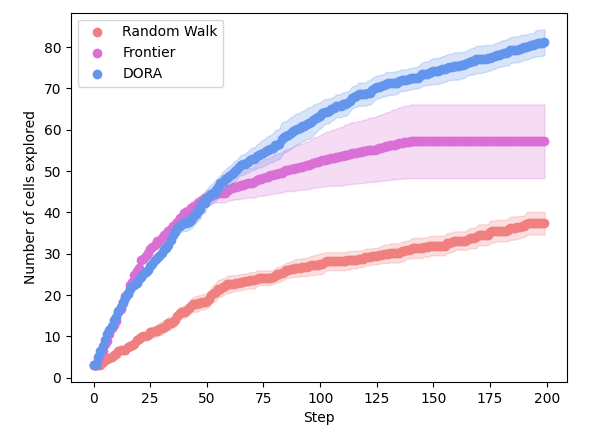}
    \caption{Performance comparison of DORA, FBE and random walk for number of explored cells over time on physical robots.}
    \label{results:cells_explored_physical}
\end{figure}

\begin{figure}[h]
    \centering
    \includegraphics[width=0.95\columnwidth]{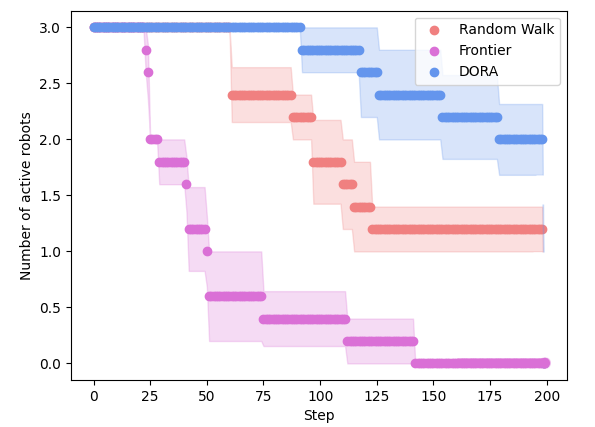}
    \caption{Performance comparison of DORA, FBE and random walk for number of active robots over time on physical robots.}
    \label{results:active_robots_physical}
\end{figure}

Figs.~\ref{results:cells_explored_physical} and
\ref{results:active_robots_physical} show an average of the results
obtained in the 5 runs of each algorithm on physical robots.

At the beginning of the exploration process, DORA and FBE perform
similarly in terms of number of cells explored as shown in
Fig. \ref{results:cells_explored_physical}. The random walk
algorithm's exploration rate is considerably smaller which can be
attributed to the fact that some of the cells of the environment are
visited multiple times: robots sometimes come back to positions that
they just had visited since their motion is determined randomly. As
time progresses, DORA starts showing better exploration results than
FBE and at the end of the runs DORA achieves a considerably better
coverage. Both FBE and DORA clearly outperform the random walk
algorithm.

In terms of robot failures, DORA outperforms both FBE and the random
walk algorithm. This is shown in
Fig. \ref{results:active_robots_physical}, where DORA exhibits a
higher level of active robots over time. When using FBE, all three
robots always fail before the end of the 200 steps run. Random walk
shows a level of active robots that is in between DORA and FBE.

The results show that while DORA and FBE initially have similar performances, as time progresses,
DORA gets better when compared to FBE. This is directly linked to the
robot failures occurring in the system. Indeed, because FBE
experiences a lot of failures, the failed robots stop exploring and, as
a result, the exploration rate decreases dramatically. In fact, FBE
always loses all its robots before the end of the experiments, hence
the robotic team completely stops discovering new cells. In contrast,
DORA keeps most of its robots active throughout the experiment and
since the workforce does not decrease to exploration rate remains
high.

\section{Conclusion}
We presented DORA Explorer, a novel lightweight risk-aware exploration
algorithm that minimizes the risk to which robots expose themselves in
order to maximize the amount of ground they will be able to cover
without failing. We expected that our exploration algorithm, which
leverages DBMs, would greatly outperform non-coordinated solutions,
and this has been the case. Indeed, it succeeded in reducing
considerably the likeliness of robot failures while keeping similar
ground coverage performance compared to other solutions proposed in
the literature. DORA also showed good scalability thanks to its low
communication costs. It also showed applicability to real world
scenarios through experiments with physical robots.

In future work, it could be interesting to allow DORA to become more
or less risk-avoiding depending on the changing needs of the
situation. For example, in a search-and-rescue scenario, an increasing
urgency to rescue victims could motivate the willingness to take more
risks as time progresses. Also, more experiments could be conducted by
testing DORA Explorer on a larger team of physical robots exploring
larger outdoor environments. Further applications of DORA could
include using the generated risk belief map to determine robots'
fitness to store data in distributed storage systems like SwarmMesh
\cite{majcherczykSwarmmesh2020}, with robots assigned to tasks in
dangerous regions being discouraged from storing sensitive
information. Additionally, in this work we considered that the risk
associated with the environment can be sensed by the robots. However,
in some scenarios, the risk cannot be directly perceived by any
sensors. In these cases, the belief map could be constructed using the
previous failures of the agents by assigning risk to areas where
failures have been detected in the past.

\bibliographystyle{IEEEtran}

\begin{thebibliography}{10}
\providecommand{\url}[1]{#1}
\csname url@samestyle\endcsname
\providecommand{\newblock}{\relax}
\providecommand{\bibinfo}[2]{#2}
\providecommand{\BIBentrySTDinterwordspacing}{\spaceskip=0pt\relax}
\providecommand{\BIBentryALTinterwordstretchfactor}{4}
\providecommand{\BIBentryALTinterwordspacing}{\spaceskip=\fontdimen2\font plus
\BIBentryALTinterwordstretchfactor\fontdimen3\font minus
  \fontdimen4\font\relax}
\providecommand{\BIBforeignlanguage}[2]{{%
\expandafter\ifx\csname l@#1\endcsname\relax
\typeout{** WARNING: IEEEtran.bst: No hyphenation pattern has been}%
\typeout{** loaded for the language `#1'. Using the pattern for}%
\typeout{** the default language instead.}%
\else
\language=\csname l@#1\endcsname
\fi
#2}}
\providecommand{\BIBdecl}{\relax}
\BIBdecl

\bibitem{matos2016multiple}
\BIBentryALTinterwordspacing
A.~Matos, A.~Martins, A.~Dias, B.~Ferreira, J.~M. Almeida, H.~Ferreira,
  G.~Amaral, A.~Figueiredo, R.~Almeida, and F.~Silva, ``Multiple robot
  operations for maritime search and rescue in eurathlon 2015 competition,'' in
  \emph{OCEANS 2016-Shanghai}.\hskip 1em plus 0.5em minus 0.4em\relax IEEE,
  2016, pp. 1--7. [Online]. Available:
  \url{https://ieeexplore.ieee.org/abstract/document/7485707}
\BIBentrySTDinterwordspacing

\bibitem{fong2005interaction}
\BIBentryALTinterwordspacing
T.~Fong and I.~Nourbakhsh, ``Interaction challenges in human-robot space
  exploration,'' \emph{Interactions}, vol.~12, no.~2, pp. 42--45, 2005.
  [Online]. Available:
  \url{https://dl.acm.org/doi/fullHtml/10.1145/1052438.1052462}
\BIBentrySTDinterwordspacing

\bibitem{burgard2005coordinated}
\BIBentryALTinterwordspacing
W.~Burgard, M.~Moors, C.~Stachniss, and F.~E. Schneider, ``Coordinated
  multi-robot exploration,'' \emph{IEEE Transactions on robotics}, vol.~21,
  no.~3, pp. 376--386, 2005. [Online]. Available:
  \url{https://ieeexplore.ieee.org/abstract/document/1435481}
\BIBentrySTDinterwordspacing

\bibitem{ramachandran2019resilience}
\BIBentryALTinterwordspacing
R.~K. Ramachandran, J.~A. Preiss, and G.~S. Sukhatme, ``Resilience by
  reconfiguration: Exploiting heterogeneity in robot teams,'' \emph{arXiv
  preprint arXiv:1903.04856}, 2019. [Online]. Available:
  \url{https://arxiv.org/abs/1903.04856}
\BIBentrySTDinterwordspacing

\bibitem{wehbe2021probabilistic}
\BIBentryALTinterwordspacing
R.~Wehbe and R.~K. Williams, ``Probabilistic resilience of dynamic multi-robot
  systems,'' \emph{IEEE Robotics and Automation Letters}, vol.~6, no.~2, pp.
  1777--1784, 2021. [Online]. Available:
  \url{https://ieeexplore.ieee.org/abstract/document/9357930}
\BIBentrySTDinterwordspacing

\bibitem{winfield2006safety}
\BIBentryALTinterwordspacing
A.~F. Winfield and J.~Nembrini, ``Safety in numbers: fault-tolerance in robot
  swarms,'' \emph{International Journal of Modelling, Identification and
  Control}, vol.~1, no.~1, pp. 30--37, 2006. [Online]. Available:
  \url{https://www.inderscienceonline.com/doi/abs/10.1504/IJMIC.2006.008645}
\BIBentrySTDinterwordspacing

\bibitem{Pinciroli:SI2012}
C.~Pinciroli, V.~Trianni, R.~O'Grady, G.~Pini, A.~Brutschy, M.~Brambilla,
  N.~Mathews, E.~Ferrante, G.~{Di Caro}, F.~Ducatelle, M.~Birattari, L.~M.
  Gambardella, and M.~Dorigo, ``{ARGoS}: a modular, parallel, multi-engine
  simulator for multi-robot systems,'' \emph{Swarm Intelligence}, vol.~6,
  no.~4, pp. 271--295, 2012.

\bibitem{amigoni2017multirobot}
\BIBentryALTinterwordspacing
F.~Amigoni, J.~Banfi, and N.~Basilico, ``Multirobot exploration of
  communication-restricted environments: A survey,'' \emph{IEEE Intelligent
  Systems}, vol.~32, no.~6, pp. 48--57, 2017. [Online]. Available:
  \url{https://ieeexplore.ieee.org/abstract/document/8267592}
\BIBentrySTDinterwordspacing

\bibitem{pinciroliTuple2016}
\BIBentryALTinterwordspacing
C.~Pinciroli, A.~Lee-Brown, and G.~Beltrame, ``A tuple space for data sharing
  in robot swarms,'' in \emph{Proceedings of the 9th EAI International
  Conference on Bio-inspired Information and Communications Technologies
  (formerly BIONETICS)}, 2016, pp. 287--294. [Online]. Available:
  \url{https://carlo.pinciroli.net/pdf/Pinciroli:BICT2015.pdf}
\BIBentrySTDinterwordspacing

\bibitem{pinciroliBuzz2016}
C.~Pinciroli and G.~Beltrame, ``Buzz: A programming language for robot
  swarms,'' \emph{IEEE Software}, vol.~33, no.~4, pp. 97--100, 2016.

\bibitem{majcherczykSwarmmesh2020}
\BIBentryALTinterwordspacing
N.~Majcherczyk and C.~Pinciroli, ``Swarmmesh: A distributed data structure for
  cooperative multi-robot applications,'' in \emph{2020 IEEE International
  Conference on Robotics and Automation (ICRA)}.\hskip 1em plus 0.5em minus
  0.4em\relax IEEE, 2020, pp. 4059--4065. [Online]. Available:
  \url{https://ieeexplore.ieee.org/abstract/document/9197403}
\BIBentrySTDinterwordspacing

\bibitem{stachnissMappingExplorationMobile2003}
C.~Stachniss and W.~Burgard, ``\BIBforeignlanguage{en}{Mapping and exploration
  with mobile robots using coverage maps},'' in
  \emph{\BIBforeignlanguage{en}{Proceedings 2003 {{IEEE}}/{{RSJ International
  Conference}} on {{Intelligent Robots}} and {{Systems}} ({{IROS}} 2003)
  ({{Cat}}. {{No}}.{{03CH37453}})}}, vol.~1.\hskip 1em plus 0.5em minus
  0.4em\relax {Las Vegas, Nevada, USA}: {IEEE}, 2003, pp. 467--472.

\bibitem{kobayashiSharingExploringInformation2002}
F.~Kobayashi, S.~Sakai, and F.~Kojima, ``Sharing of exploring information using
  belief measure for multi robot exploration,'' in \emph{2002 {{IEEE World
  Congress}} on {{Computational Intelligence}}. 2002 {{IEEE International
  Conference}} on {{Fuzzy Systems}}. {{FUZZ}}-{{IEEE}}'02. {{Proceedings}}
  ({{Cat}}. {{No}}.{{02CH37291}})}, vol.~2, May 2002, pp. 1544--1549 vol.2.

\bibitem{kobayashiDeterminationExplorationTarget2003}
------, ``Determination of exploration target based on belief measure in
  multi-robot exploration,'' in \emph{Proceedings 2003 {{IEEE International
  Symposium}} on {{Computational Intelligence}} in {{Robotics}} and
  {{Automation}}. {{Computational Intelligence}} in {{Robotics}} and
  {{Automation}} for the {{New Millennium}} ({{Cat}}. {{No}}.{{03EX694}})},
  vol.~3, Jul. 2003, pp. 1545--1550 vol.3.

\bibitem{indelmanCooperativeMultirobotBelief2018}
V.~Indelman, ``\BIBforeignlanguage{en}{Cooperative multi-robot belief space
  planning for autonomous navigation in unknown environments},''
  \emph{\BIBforeignlanguage{en}{Autonomous Robots}}, vol.~42, no.~2, pp.
  353--373, Feb. 2018.

\bibitem{hanGridWiseControlMultiAgent}
\BIBentryALTinterwordspacing
L.~Han, P.~Sun, Y.~Du, J.~Xiong, Q.~Wang, X.~Sun, H.~Liu, and T.~Zhang,
  ``Grid-wise control for multi-agent reinforcement learning in video game
  {AI},'' in \emph{Proceedings of the 36th International Conference on Machine
  Learning}, ser. Proceedings of Machine Learning Research, K.~Chaudhuri and
  R.~Salakhutdinov, Eds., vol.~97.\hskip 1em plus 0.5em minus 0.4em\relax PMLR,
  09--15 Jun 2019, pp. 2576--2585. [Online]. Available:
  \url{http://proceedings.mlr.press/v97/han19a.html}
\BIBentrySTDinterwordspacing

\bibitem{panovGridPathPlanning2018}
A.~I. Panov, K.~S. Yakovlev, and R.~Suvorov, ``\BIBforeignlanguage{en}{Grid
  {{Path Planning}} with {{Deep Reinforcement Learning}}: {{Preliminary
  Results}}},'' \emph{\BIBforeignlanguage{en}{Procedia Computer Science}}, vol.
  123, pp. 347--353, 2018.

\bibitem{undurti2010online}
\BIBentryALTinterwordspacing
A.~Undurti and J.~P. How, ``An online algorithm for constrained pomdps,'' in
  \emph{2010 IEEE International Conference on Robotics and Automation}.\hskip
  1em plus 0.5em minus 0.4em\relax IEEE, 2010, pp. 3966--3973. [Online].
  Available: \url{https://doi.org/10.1109/ROBOT.2010.5509743}
\BIBentrySTDinterwordspacing

\bibitem{thiebaux2016rao}
\BIBentryALTinterwordspacing
S.~Thi{\'e}baux, B.~Williams \emph{et~al.}, ``Rao*: An algorithm for
  chance-constrained pomdp's,'' in \emph{Proceedings of the AAAI Conference on
  Artificial Intelligence}, vol.~30, no.~1, 2016. [Online]. Available:
  \url{https://ojs.aaai.org/index.php/AAAI/article/view/10423}
\BIBentrySTDinterwordspacing

\bibitem{xiao2020robot}
\BIBentryALTinterwordspacing
X.~Xiao, J.~Dufek, and R.~R. Murphy, ``Robot risk-awareness by formal risk
  reasoning and planning,'' \emph{IEEE Robotics and Automation Letters},
  vol.~5, no.~2, pp. 2856--2863, 2020. [Online]. Available:
  \url{https://doi.org/10.1109/LRA.2020.2974434}
\BIBentrySTDinterwordspacing

\bibitem{luo2019voronoi}
\BIBentryALTinterwordspacing
W.~Luo and K.~Sycara, ``Voronoi-based coverage control with connectivity
  maintenance for robotic sensor networks,'' in \emph{2019 International
  Symposium on Multi-Robot and Multi-Agent Systems (MRS)}.\hskip 1em plus 0.5em
  minus 0.4em\relax IEEE, 2019, pp. 148--154. [Online]. Available:
  \url{https://ieeexplore.ieee.org/document/8901078}
\BIBentrySTDinterwordspacing

\bibitem{santos2019decentralized}
\BIBentryALTinterwordspacing
M.~Santos, S.~Mayya, G.~Notomista, and M.~Egerstedt, ``Decentralized
  minimum-energy coverage control for time-varying density functions,'' in
  \emph{2019 International Symposium on Multi-Robot and Multi-Agent Systems
  (MRS)}.\hskip 1em plus 0.5em minus 0.4em\relax IEEE, 2019, pp. 155--161.
  [Online]. Available: \url{https://ieeexplore.ieee.org/document/8901076}
\BIBentrySTDinterwordspacing

\bibitem{xu2019multi}
\BIBentryALTinterwordspacing
X.~Xu and Y.~Diaz-Mercado, ``Multi-robot control using coverage over
  time-varying domains,'' in \emph{2019 International Symposium on Multi-Robot
  and Multi-Agent Systems (MRS)}.\hskip 1em plus 0.5em minus 0.4em\relax IEEE,
  2019, pp. 179--181. [Online]. Available:
  \url{https://ieeexplore.ieee.org/document/8901067}
\BIBentrySTDinterwordspacing

\bibitem{yamauchi1998frontier}
\BIBentryALTinterwordspacing
B.~Yamauchi, ``Frontier-based exploration using multiple robots,'' in
  \emph{Proceedings of the second international conference on Autonomous
  agents}, 1998, pp. 47--53. [Online]. Available:
  \url{https://dl.acm.org/doi/abs/10.1145/280765.280773}
\BIBentrySTDinterwordspacing

\bibitem{wang2011frontier}
\BIBentryALTinterwordspacing
Y.~Wang, A.~Liang, and H.~Guan, ``Frontier-based multi-robot map exploration
  using particle swarm optimization,'' in \emph{2011 IEEE symposium on Swarm
  intelligence}.\hskip 1em plus 0.5em minus 0.4em\relax IEEE, 2011, pp. 1--6.
  [Online]. Available:
  \url{https://ieeexplore.ieee.org/abstract/document/5952584}
\BIBentrySTDinterwordspacing

\bibitem{topiwala2018frontier}
\BIBentryALTinterwordspacing
A.~Topiwala, P.~Inani, and A.~Kathpal, ``Frontier based exploration for
  autonomous robot,'' \emph{arXiv preprint arXiv:1806.03581}, 2018. [Online].
  Available: \url{https://arxiv.org/abs/1806.03581}
\BIBentrySTDinterwordspacing

\bibitem{dames2012decentralized}
\BIBentryALTinterwordspacing
P.~Dames, M.~Schwager, V.~Kumar, and D.~Rus, ``A decentralized control policy
  for adaptive information gathering in hazardous environments,'' in \emph{2012
  IEEE 51st IEEE Conference on Decision and Control (CDC)}.\hskip 1em plus
  0.5em minus 0.4em\relax IEEE, 2012, pp. 2807--2813. [Online]. Available:
  \url{https://ieeexplore.ieee.org/abstract/document/6426239}
\BIBentrySTDinterwordspacing

\bibitem{schwagerMultirobotControlPolicy2017}
M.~Schwager, P.~Dames, D.~Rus, and V.~Kumar, ``\BIBforeignlanguage{en}{A
  {{Multi}}-robot {{Control Policy}} for {{Information Gathering}} in the
  {{Presence}} of {{Unknown Hazards}}},'' in
  \emph{\BIBforeignlanguage{en}{Robotics {{Research}} : {{The}} 15th
  {{International Symposium ISRR}}}}, ser. Springer {{Tracts}} in {{Advanced
  Robotics}}, H.~I. Christensen and O.~Khatib, Eds.\hskip 1em plus 0.5em minus
  0.4em\relax {Cham}: {Springer International Publishing}, 2017, pp. 455--472.

\bibitem{shahriari2018lightweight}
\BIBentryALTinterwordspacing
M.~Shahriari, I.~{\v{S}}vogor, D.~St-Onge, and G.~Beltrame, ``Lightweight
  collision avoidance for resource-constrained robots,'' in \emph{2018 IEEE/RSJ
  International Conference on Intelligent Robots and Systems (IROS)}.\hskip 1em
  plus 0.5em minus 0.4em\relax IEEE, 2018, pp. 1--9. [Online]. Available:
  \url{https://ieeexplore.ieee.org/abstract/document/8593841}
\BIBentrySTDinterwordspacing

\bibitem{kteam2021kheperaiv}
\BIBentryALTinterwordspacing
{K-Team}, ``{Khepera IV},'' 2021. [Online]. Available:
  \url{https://www.k-team.com/khepera-iv}
\BIBentrySTDinterwordspacing

\end{thebibliography}

\end{document}